\documentclass[runningheads]{llncs}
\usepackage[T1]{fontenc}

\usepackage{graphicx}
\usepackage{booktabs}
\usepackage{multirow}
\usepackage{array}
\usepackage{amsmath}
\usepackage{amssymb}
\usepackage{microtype}
\usepackage[table]{xcolor}
\usepackage{hyperref}
\usepackage{url}
\usepackage{float}
\usepackage{tabularx}
\usepackage{bbding}   
\usepackage{ulem}
\newcommand{\authsup}[1]{\textsuperscript{\smash{#1}}}
\newcommand{\authorrow}[1]{\rule{0pt}{2.6ex}#1}
\begin{document}

\title{FetalAgents: A Multi-Agent System for Fetal Ultrasound Image and Video Analysis}
\titlerunning{FetalAgents: A Multi-Agent System for Fetal US}

\author{
\begin{tabular}[t]{@{}c@{}}
\authorrow{
\mbox{Xiaotian~Hu\authsup{1,*}},
\mbox{Junwei~Huang\authsup{1,*}},
\mbox{Mingxuan~Liu\authsup{1,*}},
\mbox{Kasidit~Anmahapong\authsup{1}},
}\\[-2.6pt]
\authorrow{
\mbox{Yifei~Chen\authsup{1}},
\mbox{Yitong~Luo\authsup{1}},
\mbox{Yiming~Huang\authsup{2}},
\mbox{Xuguang~Bai\authsup{1}},
\mbox{Zihan~Li\authsup{1}},
}\\[-2.6pt]
\authorrow{
\mbox{Yi~Liao\authsup{3}},
\mbox{Haibo~Qu\authsup{3}},
and \mbox{Qiyuan~Tian\authsup{1,\Envelope}}
}
\end{tabular}
}

\authorrunning{X. Hu et al.}

\institute{
\parbox[t]{\textwidth}{\centering
\textsuperscript{1} Tsinghua University, Beijing, China\\
\textsuperscript{2} University of California San Diego, La Jolla, CA, USA\\
\textsuperscript{3} West China Second University Hospital, Sichuan University, Chengdu, China\\[4pt]
\texttt{qiyuantian@tsinghua.edu.cn}
}
}

\maketitle

\begingroup
\renewcommand\thefootnote{}
\footnotetext{* These authors contributed equally.}
\footnotetext{\Envelope\ Corresponding author.}
\endgroup

\begin{abstract}

Fetal ultrasound (US) is the primary imaging modality for prenatal screening, yet its interpretation relies heavily on the expertise of the clinician. Despite advances in deep learning and foundation models, existing automated tools for fetal US analysis struggle to balance task-specific accuracy with the whole-process versatility required to support end-to-end clinical workflows. To address these limitations, we propose \textbf{\texttt{FetalAgents}}, the first multi-agent system for comprehensive fetal US analysis. Through a lightweight, agentic coordination framework, \textbf{\texttt{FetalAgents}} dynamically orchestrates specialized vision experts to maximize performance across diagnosis, measurement, and segmentation. Furthermore, \textbf{\texttt{FetalAgents}} advances beyond static image analysis by supporting end-to-end video stream summarization, where keyframes are automatically identified across multiple anatomical planes, analyzed by coordinated experts, and synthesized with patient metadata into a structured clinical report. Extensive multi-center external evaluations across eight clinical tasks demonstrate that \textbf{\texttt{FetalAgents}} consistently delivers the most robust and accurate performance when compared against specialized models and multimodal large language models (MLLMs), ultimately providing an auditable, workflow-aligned solution for fetal ultrasound analysis and reporting. Source code is available at: \url{https://github.com/huang-jw22/FetalAgents}


\end{abstract}
\section{Introduction}
Fetal ultrasound (US) is the cornerstone imaging modality for routine prenatal screening and structural anomaly assessment \cite{salomon2011routine}. However, its interpretation relies heavily on examiner expertise, and the scarcity of experienced sonographers further amplifies disparities in care quality \cite{sippel2011ultrasound}. To overcome these limitations, deep learning has been extensively explored to automate and standardize fetal US analysis \cite{fiorentino2023review}. Specifically, task-specific architectures, ranging from residual networks \cite{he2016resnet} and vision transformers \cite{dosovitskiy2021vit} for plane classification and gestational age (GA) regression, to U-Net \cite{ronneberger2015unet} variants \cite{isensee2021nnunet} and SAM-based adaptations \cite{kirillov2023sam,lin2023samus,zhou2025aopsam} for anatomical segmentation and biometry, have established highly accurate automated baselines across individual fetal US subtasks. More recently, foundation models such as FetalCLIP \cite{maani2025fetalclip} and USFM \cite{jiao2024usfm} leverage large-scale pre-training to extract versatile representations generalizable across diverse downstream clinical tasks. However, most automated models remain narrowly specialized and workflow-fragmented, typically optimized for a single subtask and deployed as standalone predictors that leave clinicians to manually coordinate tool selection and translate predictions into structured documentation \cite{fiorentino2023review,yan2025progress}.

Recent advances in agentic systems offer a promising approach to overcoming these limitations. By augmenting large language models (LLMs) with planning, memory, and tool-use capabilities \cite{hu2026landscape}, agentic systems can orchestrate specialized models to execute multi-step workflows, as demonstrated by multi-disciplinary consultation agents \cite{kim2024mdagents,chen2025mdteamgpt}, longitudinal disease-tracking systems \cite{li2025caread,DeepRare_2026}, and multimodal medical imaging frameworks \cite{fallahpour2025medrax,lyu2025wsiagents,vaidya2025medpao}. However, no existing agentic system addresses the unique demands of fetal US, where clinical reasoning spans multiple anatomical planes, tightly couples classification, segmentation, and biometry, and must handle temporal video streams beyond static images. 

To bridge this gap, we present \textbf{\texttt{FetalAgents}}, the first multi-agent system for fetal US analysis. \textbf{\texttt{FetalAgents}} orchestrates a diverse suite of domain-specific models through a lightweight, language-driven coordination framework that dynamically interprets clinical intent, identifies anatomical context, and dispatches tasks to dedicated expert modules, thereby emulating the holistic reasoning process of human sonographers. Furthermore, recognizing that real-world fetal US examination is a temporal and continuous process, \textbf{\texttt{FetalAgents}} extends beyond static image analysis to support end-to-end video stream summarization. Specifically, an automated keyframe extraction module distills continuous scans into diagnostically relevant frames, which are analyzed by coordinated experts and synthesized into structured clinical reports. Extensive experiments on multi-center external datasets confirm that \textbf{\texttt{FetalAgents}} consistently achieves the best performance across eight clinical tasks when compared with specialized vision models and MLLMs, validating its potential to standardize and automate real-world sonographic workflows.

\section{Methods}

\begin{figure}[t]
    \centering
    \includegraphics[width=\textwidth]{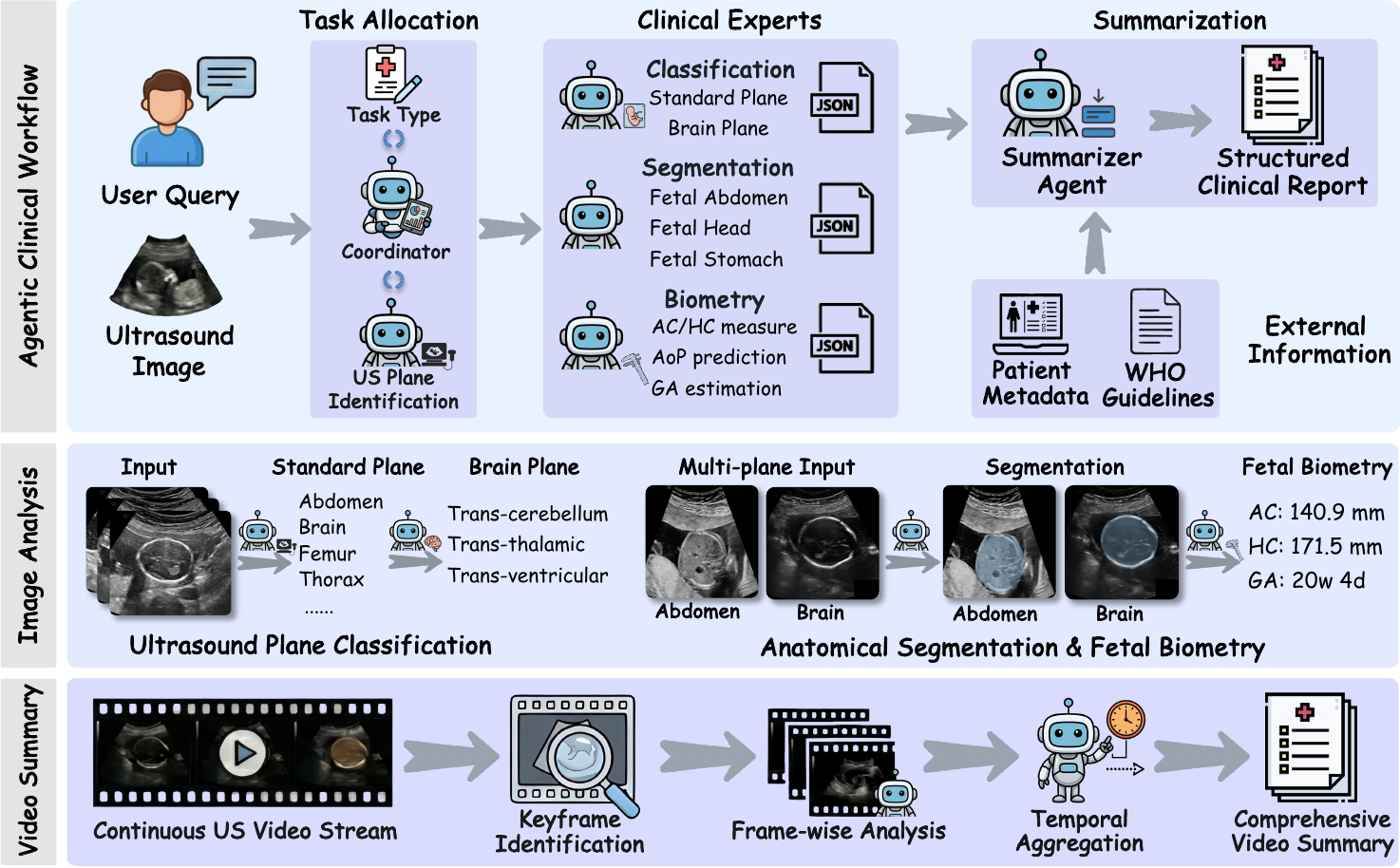}
    \caption{System diagram of FetalAgents. The framework comprises three distinct modules: (1) Agentic Clinical Workflow, where the Coordinator interprets queries and dispatches tasks; (2) Image Analysis, where specialized experts execute specific vision tasks (classification, segmentation, biometry); and (3) Video Summary, which aggregates multi-plane keyframes and temporal findings into a comprehensive report.}
    \label{fig:diagram}
\end{figure}

\subsection{Overview of \textbf{\texttt{FetalAgents}}}
\textbf{\texttt{FetalAgents}} is a collaborative, multi-agent system designed for comprehensive fetal ultrasound (US) analysis. Built upon the AutoGen \cite{wu2023autogen} framework, \textbf{\texttt{FetalAgents}} integrates specialized vision tools with reasoning LLM agents, bridging the gap between isolated point-prediction models and comprehensive, real-life clinical workflows. To provide stable and reliable fetal US analysis while minimizing hallucination risks, \textbf{\texttt{FetalAgents}} distributes distinct responsibilities across three specialized agent types: Coordinator, Experts, and Summarizer:

(1) \underline{Coordinator Agent.} Based on the GPT-5-mini \cite{openai_gpt5mini_2026}, the Coordinator acts as an intelligent central orchestrator of the entire system. Given an input US image $I$ (or a video stream $V$) and a user query $Q$, the Coordinator first classifies the query into a task type $\tau$ (specific clinical tasks or report generation), and identifies the standard ultrasound plane $c$. Based on the combined information $(Q, \tau, c)$, it rephrases the query into a structured prompt $P$, and dynamically dispatches $(I, P)$ to a subset of relevant specialized experts.

(2) \underline{Expert Agents.} Modular Expert Agents are dedicated to addressing specific clinical tasks. Each expert $\mathcal{E}_k$ wraps a set of task-specific vision models $\mathcal{M}_k = \{m_{k,1}, m_{k,2}, \dots, m_{k,N}\}$ as callable tools and applies deterministic fusion rules $\mathcal{F}_k$ to generate the most robust results $\hat{o}_k = \mathcal{F}(m_{k,1}(I), \dots, m_{k,N}(I))$ across multiple outputs. Crucially, experts communicate upstream exclusively via structured JSON formats, eliminating subjective hallucinations from intermediate steps.

(3) \underline{Summarizer Agent.} The Summarizer functions as a comprehensive report generator. It aggregates the structured outputs and tool-decision notes from the Experts to formulate a cohesive, human-readable report. By cross-referencing biometric findings from related tasks, the Summarizer acts as a final clinical safeguard before formatting the final response.

\subsection{Fetal Ultrasound Analysis}
To support diverse clinical scenarios, \textbf{\texttt{FetalAgents}} is equipped with a robust toolset that addresses three primary categories of fetal US tasks:

\underline{Plane Classification.} Accurate identification of standard fetal planes is a critical prerequisite for morphological diagnosis. To achieve this, \textbf{\texttt{FetalAgents}} integrates an ensemble of four complementary models $\mathcal{M}_{plane}$  to recognize standard fetal views and three brain sub-planes: (1) FetalCLIP \cite{maani2025fetalclip}, a vision-language foundation model capable of generating universal representations of fetal US images; (2) FU-LoRA \cite{wang2024fulora}, a Latent Diffusion Model fine-tuned via Low-Rank Adaptation (LoRA) \cite{hu2021lora}; (3) a ResNet-50 backbone initialized with RadImageNet \cite{mei2022radimagenet} weights; and (4) a Vision Transformer (ViT-B/16) \cite{dosovitskiy2021vit} initialized with ImageNet-1K \cite{deng2009imagenet} weights.

\underline{Anatomical Segmentation.} Fine-grained segmentation of anatomical structures, such as the fetal head, abdomen, and stomach, provides crucial geometric priors for downstream biometry. For fetal head segmentation, the system ensembles nnU-Net \cite{isensee2021nnunet}, the Ultrasound Foundation Model (USFM) \cite{jiao2024usfm}, and a lightweight CNN \cite{CSM}. For abdominal segmentation, we utilize FetalCLIP to generate a coarse initial mask, subsequently extracting its geometric center and bounding box to serve as input prompts for SAMUS \cite{lin2023samus}, an interactive, SAM-based US segmentation model. Finally, for the fetal stomach, we deploy a robust ensemble comprising nnU-Net, FetalCLIP, and SAMUS to ensure maximal segmentation fidelity.

\underline{Fetal Biometry.} Fetal biometry is essential for tracking fetal development and guiding intrapartum care. To automatically measure the Angle of Progression (AoP), the system employs an ensemble of AoP-SAM \cite{zhou2025aopsam} (a tailored SAM for pubic symphysis and fetal head segmentation), USFM, and an adapted UperNet \cite{Upernet} model. For precise GA estimation, we integrate FetalCLIP with dedicated regression heads fine-tuned using ResNet-50 and ConvNeXt-Tiny \cite{liu2022convnet} backbones. Routine biometric indices, including Head Circumference (HC) and Abdominal Circumference (AC), are deterministically derived from the high-fidelity masks generated by the anatomical segmentation experts using ellipse fitting.

\subsection{Comprehensive Clinical Workflow}
Ultrasound scans are inherently multi-plane and temporal. Traditional models typically operate on a single static frame and a single task, leaving major workflow steps manual and operator-dependent \cite{guo2024mmsummary}. \textbf{\texttt{FetalAgents}} mirrors real-world clinical workflows by extending beyond isolated tasks to support comprehensive image captioning and video stream summarization.

\underline{Image Caption Generation.} When presented with a broad request, \textbf{\texttt{FetalAgents}} leverages its inherent agentic intelligence to synthesize findings. The Coordinator first identifies the anatomical plane and invokes the full suite of plane-appropriate experts. The Summarizer then integrates these multi-faceted outputs $\mathcal{J}$ into a structured, comprehensive report template, ensuring all available findings are consolidated consistently. Additionally, during image caption generation, \textbf{\texttt{FetalAgents}} evaluates the biometric measurement (AC, HC) against the estimated GA by calculating the normative growth percentile $p$ using the WHO fetal growth charts \cite{WHO}. Importantly, the normative percentile doubles as an automated consistency safeguard. Measurements that deviate beyond plausible growth bounds prompt the Summarizer to invoke a reflection step over the expert outputs, dynamically replacing the outlier with a more trustworthy ensemble prediction to uphold reporting accuracy.

\underline{Video Summarization.} Manually extracting keyframes and documenting frame-wise findings from US video streams $V = \{I_1, I_2, \dots, I_T\}$ is highly labor-intensive and operator-dependent. To automate this process, we fine-tuned a 6-class keyframe identification model utilizing the FetalCLIP encoder backbone. \textbf{\texttt{Fetal\-Agents}} then deploys this model to automatically extract a subset of diagnostic keyframes $\mathcal{K} = \{I_{k_1}, \dots, I_{k_M}\} \subset V$, assigns appropriate Experts for frame-wise analysis, and finally, aggregates the temporal findings into a holistic, video-level clinical report. Through
this end-to-end video summarization pipeline,
\textbf{\texttt{FetalAgents}} effectively mirrors real-world sonographer
workflows, offering significant adaptability over existing rigid
video-summarization pipelines and seamlessly accommodating diverse
clinical scenarios without extensive workflow-specific fine-tuning.

\section{Experiments}

\subsection{Datasets and Experimental Setup}
\textbf{Training and external validation datasets.} \textbf{\texttt{FetalAgents}} was trained using multiple public fetal US datasets for different task-specific purposes, including Fetal\_PLANES\_DB \cite{FETAL_PLANES_DB} (plane classification), ACOUSLIC-AI \cite{sappia_acouslic-ai_2025} and the dataset from \cite{Stomach_training} (abdomen and stomach segmentation), PSFHS \cite{chen_psfhs_2024} (AoP prediction), HC18 \cite{HC18} (head circumference measurement), and PBF-US1 \cite{gonzalez_2024_14193949} (keyframe identification). \textbf{\texttt{FetalAgents}} and comparison methods were evaluated on several independent external datasets to assess out-of-distribution performance, including multi-center African data \cite{Africa,African_HC} (standard plane classification, HC measurement), datasets from \cite{Fetal_planes_and_organs} (brain plane classification, stomach segmentation), \cite{Abdomen_OOD} (abdomen segmentation), \cite{lu_jnu-ifm_2022} (AoP prediction), and a 511-case subset of HC18 \cite{HC18} (GA estimation).

\begin{table}[!htbp]
\centering
\caption{Quantitative evaluation on external datasets across eight clinical
tasks grouped into three categories. \textbf{Bold}: best. \underline{Underline}: runner-up.}
\label{tab:results}
{%
\fontsize{8}{9}\selectfont
\setlength{\tabcolsep}{5pt}
\renewcommand{\arraystretch}{0.85}
\begin{tabular}{@{}lcccccc@{}}
\toprule

\multicolumn{7}{c}{\cellcolor{gray!30}\textbf{I.\quad Plane Classification}} \\
\midrule
\textbf{Method} & \textbf{Accuracy} & \textbf{Precision} & \textbf{Recall}
  & \textbf{F1-score} & \textbf{Cohen's $\kappa$} & \textbf{AUROC} \\
\midrule
\multicolumn{7}{l}{\cellcolor{gray!10}\textit{Standard Plane Classif.\ (Ext.\ Val.: 233 cases)}} \\
\addlinespace[3pt]
FetalCLIP\cite{maani2025fetalclip}      & \underline{0.914} & \underline{0.942} & \underline{0.915}
               & \underline{0.927} & \underline{0.887} & 99.7\% \\
FU-LoRA\cite{wang2024fulora}        & 0.764 & \textbf{0.977} & 0.759
               & 0.847 & 0.717 & \underline{98.0\%} \\
GPT-5-mini\cite{openai_gpt5mini_2026}     & 0.605 & 0.533 & 0.548
               & 0.523 & 0.464 & 0.816 \\
Gemini-3-flash\cite{gemini3flash} & 0.880 & 0.894 & 0.843
               & 0.856 & 0.857 & 0.961 \\
MedGemma\cite{sellergren2025medgemmatechnicalreport}       & 0.708 & 0.831 & 0.649
               & 0.617 & 0.597 & N/A \\
\textbf{\textbf{\texttt{FetalAgents}}}  & \textbf{0.927} & 0.937 & \textbf{0.927}
               & \textbf{0.931} & \textbf{0.903} & \textbf{99.7\%} \\
\addlinespace[3pt]
\multicolumn{7}{l}{\cellcolor{gray!10}\textit{Brain Plane Classif.\ (Ext.\ Val.: 569 cases)}} \\
\addlinespace[3pt]
FetalCLIP\cite{maani2025fetalclip}      & \underline{0.893} & \underline{0.880} & \underline{0.911}
               & \underline{0.883} & \underline{0.830} & \textbf{98.6\%} \\
ResNet\cite{he2016resnet,mei2022radimagenet}         & 0.824 & 0.798 & 0.853
               & 0.810 & 0.727 & 94.7\% \\
ViT-16\cite{dosovitskiy2021vit,deng2009imagenet}         & 0.881 & 0.864 & 0.896
               & 0.874 & 0.810 & \underline{96.5\%} \\
GPT-5-mini\cite{openai_gpt5mini_2026}     & 0.689 & 0.644 & 0.632
               & 0.638 & 0.518 & 0.789 \\
Gemini-3-flash\cite{gemini3flash} & 0.636 & 0.657 & 0.661
               & 0.606 & 0.473 & 0.755 \\
MedGemma\cite{sellergren2025medgemmatechnicalreport}       & 0.691 & 0.723 & 0.693
               & 0.684 & 0.509 & N/A \\
\textbf{\textbf{\texttt{FetalAgents}}}  & \textbf{0.909} & \textbf{0.884} & \textbf{0.925}
               & \textbf{0.896} & \textbf{0.856} & 96.3\% \\
\midrule

\multicolumn{7}{c}{\cellcolor{gray!30}\textbf{II.\quad Anatomical Segmentation}} \\
\midrule
\textbf{Method} & \textbf{DSC} & \textbf{IoU} & \textbf{HD95}
  & \textbf{ASSD} & \textbf{PPV} & \textbf{Sens} \\
\midrule
\multicolumn{7}{l}{\cellcolor{gray!10}\textit{Abdomen Seg.\ (Ext.\ Val.: 187 cases)}} \\
\addlinespace[3pt]
FetalCLIP\cite{maani2025fetalclip}      & \underline{0.902} & \underline{0.834} & \underline{4.20}
               & \underline{0.447} & \underline{0.891} & \underline{0.940} \\
nnUNet\cite{isensee2021nnunet}         & 0.848 & 0.748 & 129.2
               & 15.20 & 0.809 & 0.912 \\
\textbf{\textbf{\texttt{FetalAgents}}}  & \textbf{0.937} & \textbf{0.937} & \textbf{2.32}
               & \textbf{0.202} & \textbf{0.922} & \textbf{0.955} \\
\addlinespace[3pt]
\multicolumn{7}{l}{\cellcolor{gray!10}\textit{Stomach Seg.\ (Ext.\ Val.: 253 cases)}} \\
\addlinespace[3pt]
FetalCLIP\cite{maani2025fetalclip}      & 0.771 & 0.653 & 52.56
               & 8.62 & 0.829 & 0.764 \\
SAMUS\cite{lin2023samus}          & 0.826 & 0.730 & 14.07
               & 5.68 & 0.797 & \underline{0.893} \\
nnUNet\cite{isensee2021nnunet}         & \underline{0.847} & \underline{0.754} & \underline{11.30}
               & \textbf{1.95} & \underline{0.841} & 0.891 \\
\textbf{\textbf{\texttt{FetalAgents}}}  & \textbf{0.859} & \textbf{0.767} & \textbf{10.64}
               & \underline{3.62} & \textbf{0.844} & \textbf{0.896} \\
\midrule

\multicolumn{7}{c}{\cellcolor{gray!30}\textbf{III.\quad Fetal Biometry}} \\
\midrule
\textbf{Method} & \textbf{DSC} & \textbf{HD95} & \textbf{MAE}
  & \textbf{MdAE} & \textbf{MdRAE} & \textbf{Acc@5\%} \\
\midrule
\multicolumn{7}{l}{\cellcolor{gray!10}\textit{AoP Estimation (Ext.\ Val.: 64 cases)}} \\
\addlinespace[3pt]
AoP-SAM\cite{zhou2025aopsam}       & \underline{0.927} & \underline{38.0} & \underline{5.59$^\circ$}
               & \underline{4.86$^\circ$} & \underline{3.54\%} & \underline{65.6\%} \\
USFM\cite{jiao2024usfm}           & 0.870 & 68.3 & 10.03$^\circ$
               & 8.16$^\circ$ & 6.46\% & 43.8\% \\
UperNet\cite{Upernet}        & 0.913 & 42.9 & 6.74$^\circ$
               & 6.31$^\circ$ & 4.77\% & 53.1\% \\
\textbf{\textbf{\texttt{FetalAgents}}}  & \textbf{0.928} & \textbf{37.2} & \textbf{5.44$^\circ$}
               & \textbf{4.68$^\circ$} & \textbf{3.50\%} & \textbf{67.2\%} \\
\addlinespace[3pt]
\multicolumn{7}{l}{\cellcolor{gray!10}\textit{AC Measurement (Ext.\ Val.: 187 cases)}} \\
\addlinespace[3pt]
FetalCLIP\cite{maani2025fetalclip}      & \underline{0.902} & \underline{4.20} & \underline{9.94\,mm}
               & \underline{6.59\,mm} & \underline{3.77\%} & \underline{65.6\%} \\
nnUNet\cite{isensee2021nnunet}         & 0.848 & 129.2 & 11.7\,mm
               & 8.80\,mm & 5.28\% & 47.1\% \\
\textbf{\textbf{\texttt{FetalAgents}}}  & \textbf{0.937} & \textbf{2.32} & \textbf{5.53\,mm}
               & \textbf{4.29\,mm} & \textbf{2.34\%} & \textbf{84.0\%} \\
\midrule
\textbf{Method} & \textbf{DSC} & \textbf{HD95} & \textbf{MRAE}
  & \textbf{MdRAE} & \textbf{P95\,RAE} & \textbf{Acc@5\%} \\
\midrule
\multicolumn{7}{l}{\cellcolor{gray!10}\textit{HC Measurement (Ext.\ Val.: 75 cases)}} \\
\addlinespace[3pt]
CSM\cite{CSM}            & 0.956 & \underline{1.92} & 2.05\%
               & 1.48\% & 5.75\% & 89.3\% \\
USFM\cite{jiao2024usfm}           & 0.924 & 11.54 & 3.21\%
               & 2.06\% & 13.0\% & 86.7\% \\
nnUNet\cite{isensee2021nnunet}         & \textbf{0.964} & \textbf{1.72} & \underline{1.57\%}
               & \underline{1.21\%} & \underline{3.96\%} & \textbf{98.7\%} \\
\textbf{\textbf{\texttt{FetalAgents}}}  & \underline{0.961} & 1.93 & \textbf{1.40\%}
               & \textbf{1.12\%} & \textbf{3.75\%} & \underline{96.0\%} \\
\midrule
\textbf{Method} & \textbf{Valid.} & \textbf{MAE} & \textbf{MdAE}
  & \textbf{MRAE} & \textbf{P95\,AE} & \textbf{Acc@5\%} \\
\midrule
\multicolumn{7}{l}{\cellcolor{gray!10}\textit{GA Prediction (Ext.\ Val.: 511 cases)}} \\
\addlinespace[3pt]
ResNet\cite{he2016resnet,mei2022radimagenet}    & 71.62\% & 1.73w & 1.08w
               & 6.74\% & 5.41w & 52.1\% \\
FetalCLIP\cite{maani2025fetalclip}      & \underline{82.97\%} & \underline{1.32w} & \underline{0.72w}
               & \underline{5.13\%} & \underline{4.69w} & \textbf{68.9\%} \\
ConvNext\cite{liu2022convnet}       & 74.17\% & 1.53w & 0.96w
               & 5.95\% & 5.19w & 56.75\% \\
\textbf{\textbf{\texttt{FetalAgents}}}  & \textbf{84.93\%} & \textbf{1.23w} & \textbf{0.66w}
               & \textbf{4.70\%} & \textbf{4.51w} & \underline{68.7\%} \\
\bottomrule
\end{tabular}%
}%
\end{table}

\textbf{Baselines and Metrics.}  \textbf{\texttt{FetalAgents}} was compared against specialized vision models \cite{wang2024fulora,he2016resnet,dosovitskiy2021vit,isensee2021nnunet,zhou2025aopsam,Upernet,CSM,liu2022convnet}, US-specific foundation models \cite{maani2025fetalclip,jiao2024usfm}, general MLLMs\cite{openai_gpt5mini_2026,gemini3flash}, and medical MLLMs\cite{sellergren2025medgemmatechnicalreport}. For plane classification, we report Accuracy, Precision, Recall, F1-score, Cohen's $\kappa$, and AUROC. For anatomical segmentation, we adopt the Dice Similarity
Coefficient (DSC), Intersection over Union (IoU), 95th-percentile Hausdorff Distance (HD95), Average Symmetric Surface Distance (ASSD), Positive Predictive Value (PPV), and Sensitivity (Sens). For fetal biometry, segmentation quality is again assessed with DSC and HD95, while measurement accuracy is quantified by Mean Absolute Error (MAE), Median Absolute Error (MdAE), Mean/Median Relative Absolute Error (MRAE/MdRAE), 95th-percentile Absolute/Relative Error (P95\,AE/P95\,RAE),
and the percentage of predictions within 5\% of the ground truth (Acc@5\%). For gestational age (GA) prediction in particular, following~\cite{maani2025fetalclip}, we introduce an additional \emph{Validity Rate} metric, which tests whether the ground-truth HC falls within the 2.5th--97.5th percentile range of the predicted GA on the WHO fetal growth charts~\cite{WHO}, thereby providing a clinically grounded sanity check on the plausibility of the estimated GA.

\begin{figure}[t]
    \centering
    \includegraphics[width=\textwidth]{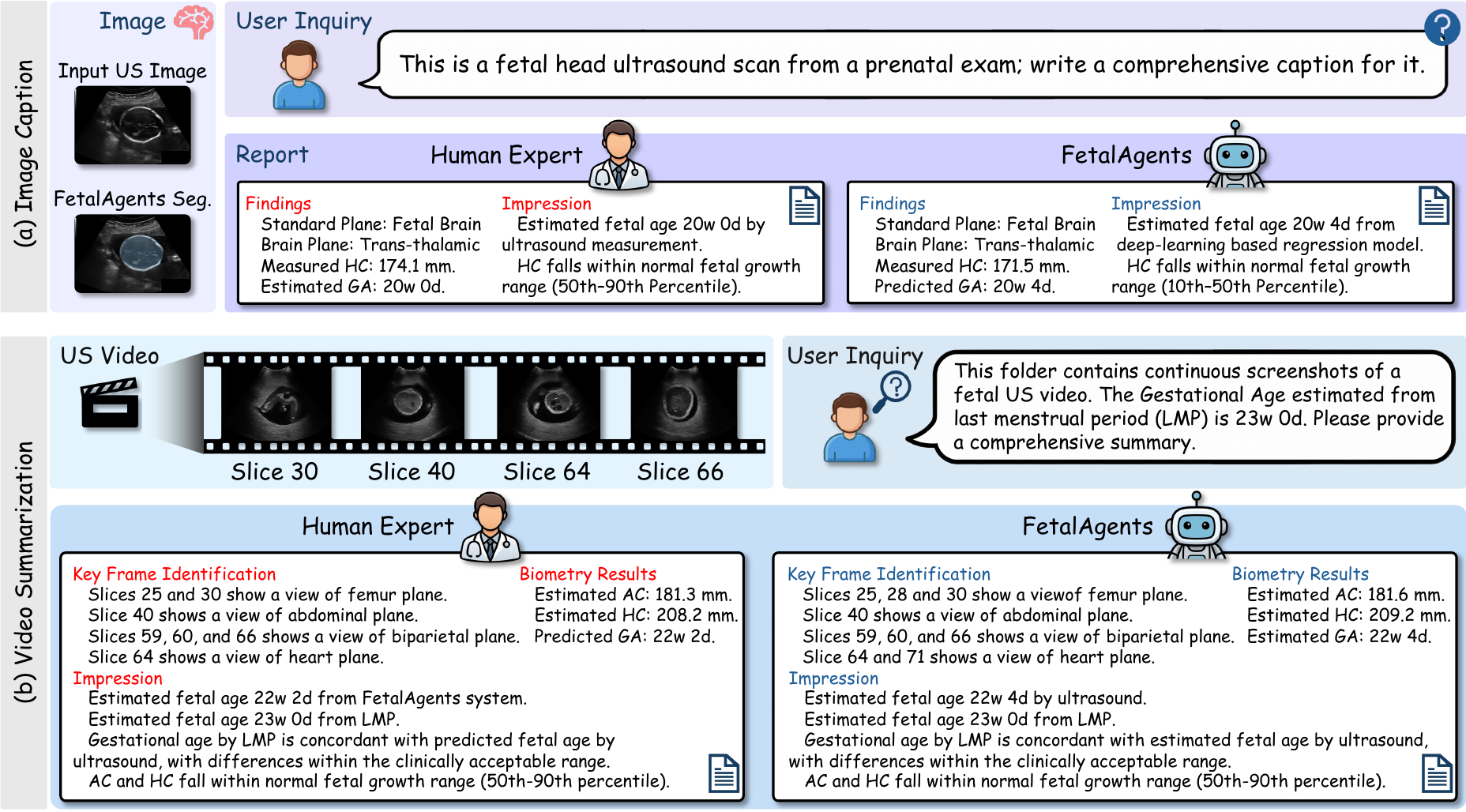}
    \caption{Qualitative evaluation of clinical reporting. (a) Image Caption Generation: FetalAgents correctly identifies the standard plane and performs biometry to generate a report concordant with human experts. (b) Video Summarization: The system autonomously extracts multi-plane keyframes from a continuous video stream and cross-references ultrasound findings with patient metadata (LMP) for clinical consistency.}
    \label{fig:demo}
\end{figure}

\subsection{Quantitative Results}
\textbf{\texttt{FetalAgents}} consistently achieves the most accurate and stable performance across 8 clinical tasks on external validation (Table~\ref{tab:results}), confirming the efficacy of agentic orchestration and dynamic model ensembling. Specifically, (1) \underline{Plane Classification:} Generic and medical MLLMs struggled significantly with the acoustically noisy US modality, with GPT-5-mini achieving only 0.605 accuracy on standard planes. By dynamically routing images to specialized visual backbones, \textbf{\texttt{FetalAgents}} outperforms the strongest standalone model FetalCLIP, improving accuracy from 0.914 to 0.927 and Cohen's $\kappa$ from 0.887 to 0.903 on standard plane classification. On the more challenging brain sub-plane task, \textbf{\texttt{FetalAgents}} achieved F1-score 0.896 and $\kappa$ 0.856, improving over FetalCLIP by 1.3\% and 2.6\%, respectively. (2) \underline{Anatomical Segmentation:} Our fusion rules effectively correct boundary errors typical of monolithic models. On abdomen segmentation, \textbf{\texttt{FetalAgents}} attained DSC 0.937 and HD95 2.32, improving over the runner-up FetalCLIP by 3.5\% in DSC and 44.8\% in HD95. For stomach segmentation, \textbf{\texttt{FetalAgents}} reached DSC 0.859 and HD95 10.64, surpassing nnUNet by 1.2\% and 5.8\%, respectively. (3) \underline{Fetal Biometry:} \textbf{\texttt{FetalAgents}} achieves the lowest errors across all biometric tasks. For AC measurement, it reduced MAE from 9.94\,mm (FetalCLIP) to 5.53\,mm and raised Acc@5\% from 65.6\% to 84.0\%. For HC measurement, it achieved MRAE 1.40\% and P95\,RAE 3.75\%, outperforming nnUNet by 10.8\% and 5.3\% in relative reduction. On GA prediction, \textbf{\texttt{FetalAgents}} attained the highest validity rate of 84.93\% and the lowest MAE of 1.23\,w, improving over FetalCLIP by 2.0 percentage points and 6.8\%, respectively. For AoP estimation, it achieved MAE 5.44$^\circ$ and Acc@5\% 67.2\%, consistently outperforming all standalone models.

\subsection{Image Captioning and Video Summarization Results}
To demonstrate the end-to-end clinical utility of \textbf{\texttt{FetalAgents}} beyond isolated metrics, we qualitatively evaluated its report generation capabilities. (1) \uline{Image Caption Generation:} Given a broad inquiry and a single US image, \textbf{\texttt{FetalAgents}} correctly identified the trans-thalamic plane, coordinated relevant experts for biometry and GA estimation, and synthesized all outputs into a structured clinical report (Fig.~\ref{fig:demo} (a)). (2) \uline{Video Summarization:} Given a continuous multi-plane video stream with patient metadata, \textbf{\texttt{FetalAgents}} autonomously extracted keyframes across diverse anatomies, aggregated frame-wise biometrics, and cross-referenced the derived GA against the LMP-derived GA to confirm clinical consistency (Fig.~\ref{fig:demo} (b)), closely mirroring real-world sonographer workflows.

\section{Conclusion}
This paper introduced \textbf{\texttt{FetalAgents}}, the first multi-agent system that bridges the gap between isolated point-prediction models and the complex realities of fetal ultrasound workflows. By leveraging inherent intelligence and tool-use capabilities of LLMs within an agentic framework, our system dynamically orchestrates diverse specialized vision backbones to conduct accurate fetal US analysis. \textbf{\texttt{FetalAgents}} successfully advances beyond static image interpretation by automating the labor-intensive process of continuous US video summarization, generating structured, expert-level clinical reports automatically. Extensive external evaluations confirm that \textbf{\texttt{FetalAgents}} consistently outperforms state-of-the-art standalone vision models and general-purpose MLLMs across a wide array of clinical tasks. Future work will focus on integrating broader longitudinal patient histories and evaluating the system in prospective clinical settings to further validate its impact.



\bibliographystyle{splncs04}
\bibliography{mybibliography}
\end{document}